\documentclass[sigconf,balance]{acmart}

\usepackage{enumitem}

\copyrightyear{2023}
\acmYear{2023}
\setcopyright{rightsretained}
\acmConference[SIGSIM-PADS '23]{ACM SIGSIM Conference on Principles of Advanced Discrete Simulation}{June 21--23, 2023}{Orlando, FL, USA}
\acmBooktitle{ACM SIGSIM Conference on Principles of Advanced Discrete Simulation (SIGSIM-PADS '23), June 21--23, 2023, Orlando, FL, USA}
\acmDOI{10.1145/3573900.3593631}
\acmISBN{979-8-4007-0030-9/23/06}

\begin{document}

\title{Autonomous Agent for Beyond Visual Range Air Combat: A Deep Reinforcement Learning Approach}

\author{Joao~P.~A.~Dantas}
\email{dantasjpad@fab.mil.br}
\orcid{0000-0003-0300-8027} 
\affiliation{%
  \institution{Institute for Advanced Studies}
  \city{Sao Jose dos Campos}
  \country{Brazil}
}

\author{Marcos~R.~O.~A.~Maximo}
\email{mmaximo@ita.br}
\orcid{0000-0003-2944-4476} 
\affiliation{%
  \institution{Aeronautics Institute of Technology}
  \city{Sao Jose dos Campos}
  \country{Brazil}
}

\author{Takashi~Yoneyama}
\email{takashi@ita.br}
\orcid{0000-0001-5375-1076} 
\affiliation{%
  \institution{Aeronautics Institute of Technology}
  \city{Sao Jose dos Campos}
  \country{Brazil}
}

\begin{abstract}
This work contributes to developing an agent based on deep reinforcement learning capable of acting in a beyond visual range (BVR) air combat simulation environment. The paper presents an overview of building an agent representing a high-performance fighter aircraft that can learn and improve its role in BVR combat over time based on rewards calculated using operational metrics. Also, through self-play experiments, it expects to generate new air combat tactics never seen before. Finally, we hope to examine a real pilot's ability, using virtual simulation, to interact in the same environment with the trained agent and compare their performances. This research will contribute to the air combat training context by developing agents that can interact with real pilots to improve their performances in air defense missions. 

\end{abstract}



\maketitle

\section{Introduction}
Air combat is a complex and dynamic scenario involving highly skilled pilots making split-second decisions to outmaneuver their opponents~\cite{dantas2022machine}. Beyond visual range (BVR) air combat, in particular, presents unique challenges and opportunities, as it involves engagements taking place at ranges beyond the pilot's ability to see the enemy aircraft~\cite{higby2005promise}. While modern air combat may still be within visual range (WVR), the combat usually begins in BVR. This phase is frequently the most critical as it can provide advantages and drawbacks for succeeding phases~\cite{dantas2018apoio}. The fundamental difficulty for pilots in the fight is maneuver planning, which reflects both sides' tactical decision-making capacity and determines success or failure~\cite{hu2021application}. Constructive computer simulations can mimic the most diverse BVR combat situations to investigate the effects of new combat tactics, sensors, and armaments~\cite{dantas2022supervised}. One of the main challenges of constructive BVR combat simulation is to simulate the complex behaviors of a pilot in all phases of combat. A pilot can perform decision-making processes such as adapting to new combat situations and conducting collective tactics with pilots from other aircraft to employ an engagement~\cite{dantas2021engagement}, or launching a missile at the proper instant~\cite{dantas2021weapon}.

The presented work proposes a model to create an autonomous agent utilizing deep reinforcement learning (DRL) to operate in a BVR air combat simulation environment. By learning from operational metrics, the agent will represent a high-performance fighter aircraft that can enhance its capabilities over time. Furthermore, we expected the generation of novel air combat tactics through self-play experiments. The ultimate goal is to allow real pilots to engage with the trained agent in the same environment using virtual simulation and compare their performances. The framework's primary objectives are to design a BVR agent that can learn all the tactics associated with this air combat mode, enhance its performance through self-play techniques, and outperform a real pilot in the context of BVR air combat.

\section{Related Work}

Several recent studies have explored DRL algorithms for autonomous decision-making in BVR air combat scenarios for different applications: generation of air combat tactics~\cite{piao2020beyond}, maneuver planning~\cite{hu2021application, zhang2022maneuver, fan2022air, zhang2022autonomous}, and multi-UAV cooperative decision-making methods~\cite{liu2022multi, hu2022autonomous}. These works demonstrate the potential of DRL-based approaches for decision-making in BVR air combat scenarios. While these approaches have shown promising results, there is still much work to do in developing more robust and sufficient algorithms and evaluating the feasibility of these methods in real-world applications. In contrast to these studies, our work focuses on applying DRL techniques to BVR air combat using a high-fidelity simulation environment. Besides, to the best of the authors' knowledge, no study has yet explored the use of DRL, combined with self-play techniques, to train a high-performance agent that can interact with a real pilot in the same simulation environment.

\section{Proposed Model}

The Rainbow algorithm~\cite{hessel2018rainbow} is a state-of-the-art DRL technique that combines several improvements to the Deep Q-Network (DQN)~\cite{mnih2015human} algorithm to achieve better results. We will use this algorithm to train our autonomous BVR air combat agent. The Rainbow algorithm includes some improvements over the DQN algorithm, including prioritized experience replay, distributional reinforcement learning, and dueling networks. Prioritized experience replay~\cite{schaul2015prioritized} allows the agent to prioritize specific transitions in the replay buffer based on their importance for learning. At the same time, distributional reinforcement learning~\cite{bellemare2017distributional} estimates the distribution of the discounted return instead of just the expected value. Finally, dueling networks~\cite{wang2015dueling} separate the estimation of the state value and the advantage function to improve the learning process.

We will use the Aerospace Simulation Environment (\emph{Ambiente de Simulação Aeroespacial -- ASA} in Portuguese) as the simulation platform to train and evaluate our agent. ASA is a custom-made, high-fidelity, object-oriented simulation framework developed mainly in C++ that enables the modeling and simulation of military operational scenarios to support the development of tactics and procedures in the aerospace context for the Brazilian Air Force~\cite{dantas2022asa}.

\begin{figure}[htb]
\centering
\includegraphics[width=0.49\textwidth]{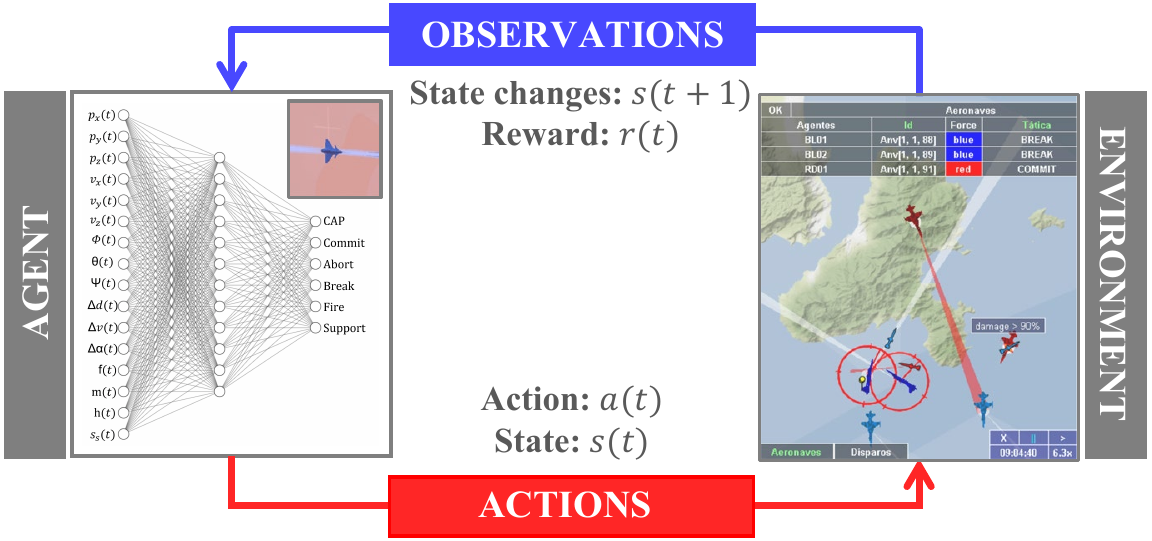}
\caption{Agent-environment interaction: $a_t$, $s_t$, and $r_t$ denote action, state, and reward at time step $t$, with $s_{t+1}$ given by the environment for the next iteraction.}
\label{fig1}
\end{figure}

To manage the agent's actions $a(t)$, we will consider the following possible tactics: Combat Air Patrol (CAP), Commit, Abort, Break, Fire, and Support. CAP involves flying around a particular location in a specific pattern; Commit means to engage a detected target; Abort is a tactic of moving away from a threat; Break is a sudden defensive maneuver when the radar detects a missile; Fire is launching a missile at a target; and Support is the tactic to aid the missile's guidance until its seeker become activate~\cite{dantas2021engagement, dantas2021weapon}. The agent's state $s(t)$ depends on the independent motion variables of the agent, such as position $[p_x(t), p_y(t), p_z(t)]$, velocity $[v_x(t), v_y(t), v_z(t)]$, and orientation regarding roll $\phi(t)$, pitch $\theta(t)$, and yaw $\psi(t)$. Besides, the state includes comparative factors between the agent and the nearest detected target, such as relative distance $\Delta d(t)$, relative speed $\Delta v(t)$, and relative angle $\Delta \alpha(t)$. Finally, the last part of the agent's state is composed of the remaining fuel $f(t)$, the remaining missiles $m(t)$, the health condition $h(t)$, and the status of the agent's sensors $s_s(t)$. During training, the Rainbow algorithm will update the neural network based on the rewards received by the agent $r(t)$, calculated by the Defensive Counter Air (DCA) Index~\cite{dantas2021engagement}, a probability of success for BVR combat on DCA missions whose objective is to establish a CAP. Figure~\ref{fig1} illustrates the agent's interaction with the environment.
 
\balance

\section{Conclusions}

This research aims to improve air combat training by developing unmanned combat aerial vehicles (UCAVs) to interact with pilots and enhance fighter performance in air defense missions. We will release the source code for the general architecture to encourage the development of diverse applications using the same simulation platform. The technology developed in this work has the potential to serve as a simulation-as-a-service (SimaaS) tool to meet various simulation needs in the defense and aerospace industries.

\bibliographystyle{ACM-Reference-Format}
\bibliography{ref}


\begin{thebibliography}{19}


\ifx \showCODEN    \undefined \def \showCODEN     #1{\unskip}     \fi
\ifx \showDOI      \undefined \def \showDOI       #1{#1}\fi
\ifx \showISBNx    \undefined \def \showISBNx     #1{\unskip}     \fi
\ifx \showISBNxiii \undefined \def \showISBNxiii  #1{\unskip}     \fi
\ifx \showISSN     \undefined \def \showISSN      #1{\unskip}     \fi
\ifx \showLCCN     \undefined \def \showLCCN      #1{\unskip}     \fi
\ifx \shownote     \undefined \def \shownote      #1{#1}          \fi
\ifx \showarticletitle \undefined \def \showarticletitle #1{#1}   \fi
\ifx \showURL      \undefined \def \showURL       {\relax}        \fi
\providecommand\bibfield[2]{#2}
\providecommand\bibinfo[2]{#2}
\providecommand\natexlab[1]{#1}
\providecommand\showeprint[2][]{arXiv:#2}

\bibitem[Bellemare et~al\mbox{.}(2017)]%
        {bellemare2017distributional}
\bibfield{author}{\bibinfo{person}{Marc~G Bellemare}, \bibinfo{person}{Will
  Dabney}, {and} \bibinfo{person}{R{\'e}mi Munos}.}
  \bibinfo{year}{2017}\natexlab{}.
\newblock \showarticletitle{A distributional perspective on reinforcement
  learning}. In \bibinfo{booktitle}{\emph{International Conference on Machine
  Learning}}. PMLR, \bibinfo{pages}{449--458}.
\newblock


\bibitem[Dantas(2018)]%
        {dantas2018apoio}
\bibfield{author}{\bibinfo{person}{Joao P.~A. Dantas}.}
  \bibinfo{year}{2018}\natexlab{}.
\newblock \emph{\bibinfo{title}{{Apoio \`{a} Decis\~{a}o para o Combate
  A\'{e}reo Al\'{e}m do Alcance Visual: Uma Abordagem por Redes Neurais
  Artificiais.}}}
\newblock {Master's Thesis}. \bibinfo{school}{Instituto Tecnol\'{o}gico de
  Aeron\'{a}utica}, \bibinfo{address}{S\~{a}o Jos\'{e} dos Campos, SP, Brazil}.
\newblock


\bibitem[Dantas et~al\mbox{.}(2021a)]%
        {dantas2021engagement}
\bibfield{author}{\bibinfo{person}{Joao P.~A. Dantas},
  \bibinfo{person}{Andre~N. Costa}, \bibinfo{person}{Diego Geraldo},
  \bibinfo{person}{Marcos R. O.~A. Maximo}, {and} \bibinfo{person}{Takashi
  Yoneyama}.} \bibinfo{year}{2021}\natexlab{a}.
\newblock \showarticletitle{Engagement Decision Support for Beyond Visual Range
  Air Combat}. In \bibinfo{booktitle}{\emph{Proceedings of the 2021 Latin
  American Robotics Symposium, 2021 Brazilian Symposium on Robotics, and 2021
  Workshop on Robotics in Education}}. \bibinfo{pages}{October
  11\textsuperscript{th}--15\textsuperscript{th}, 96--101}.
\newblock


\bibitem[Dantas et~al\mbox{.}(2021b)]%
        {dantas2021weapon}
\bibfield{author}{\bibinfo{person}{Joao P.~A. Dantas},
  \bibinfo{person}{Andre~N. Costa}, \bibinfo{person}{Diego Geraldo},
  \bibinfo{person}{Marcos R. O.~A. Maximo}, {and} \bibinfo{person}{Takashi
  Yoneyama}.} \bibinfo{year}{2021}\natexlab{b}.
\newblock \showarticletitle{Weapon Engagement Zone Maximum Launch Range
  Estimation Using a Deep Neural Network}. In
  \bibinfo{booktitle}{\emph{Intelligent Systems}},
  \bibfield{editor}{\bibinfo{person}{Andr{\'e} Britto} {and}
  \bibinfo{person}{Karina Valdivia~Delgado}} (Eds.).
  \bibinfo{publisher}{Springer}, \bibinfo{address}{Cham},
  \bibinfo{pages}{193--207}.
\newblock
\showISBNx{978-3-030-91699-2}


\bibitem[Dantas et~al\mbox{.}(2022a)]%
        {dantas2022asa}
\bibfield{author}{\bibinfo{person}{Joao P.~A Dantas}, \bibinfo{person}{Andre~N.
  Costa}, \bibinfo{person}{Vitor C.~F. Gomes}, \bibinfo{person}{Andre~R.
  Kuroswiski}, \bibinfo{person}{Felipe L.~L. Medeiros}, {and}
  \bibinfo{person}{Diego Geraldo}.} \bibinfo{year}{2022}\natexlab{a}.
\newblock \showarticletitle{{ASA: A Simulation Environment for Evaluating
  Military Operational Scenarios}}. In \bibinfo{booktitle}{\emph{Proceedings of
  the 20\textsuperscript{th} International Conference on Scientific
  Computing}}. \bibinfo{pages}{25\textsuperscript{th}--28\textsuperscript{th},
  Las Vegas, NV, USA}.
\newblock


\bibitem[Dantas et~al\mbox{.}(2022b)]%
        {dantas2022supervised}
\bibfield{author}{\bibinfo{person}{Joao P.~A. Dantas},
  \bibinfo{person}{Andre~N. Costa}, \bibinfo{person}{Felipe L.~L. Medeiros},
  \bibinfo{person}{Diego Geraldo}, \bibinfo{person}{Marcos R. O.~A. Maximo},
  {and} \bibinfo{person}{Takashi Yoneyama}.} \bibinfo{year}{2022}\natexlab{b}.
\newblock \showarticletitle{{Supervised Machine Learning for Effective Missile
  Launch Based on Beyond Visual Range Air Combat Simulations}}. In
  \bibinfo{booktitle}{\emph{Proceedings of the Winter Simulation Conference}}
  (Singapore) \emph{(\bibinfo{series}{WSC '22})}.
\newblock


\bibitem[Dantas et~al\mbox{.}(2022c)]%
        {dantas2022machine}
\bibfield{author}{\bibinfo{person}{Joao P.~A. Dantas}, \bibinfo{person}{Marcos
  R. O.~A. Maximo}, \bibinfo{person}{Andre~N. Costa}, \bibinfo{person}{Diego
  Geraldo}, {and} \bibinfo{person}{Takashi Yoneyama}.}
  \bibinfo{year}{2022}\natexlab{c}.
\newblock \showarticletitle{{Machine Learning to Improve Situational Awareness
  in Beyond Visual Range Air Combat}}.
\newblock \bibinfo{journal}{\emph{IEEE Latin America Transactions}}
  \bibinfo{volume}{20}, \bibinfo{number}{8} (\bibinfo{year}{2022}).
\newblock
\urldef\tempurl%
\url{https://latamt.ieeer9.org/index.php/transactions/article/view/6530}
\showURL{%
\tempurl}


\bibitem[Fan et~al\mbox{.}(2022)]%
        {fan2022air}
\bibfield{author}{\bibinfo{person}{Zihao Fan}, \bibinfo{person}{Yang Xu},
  \bibinfo{person}{Yuhang Kang}, {and} \bibinfo{person}{Delin Luo}.}
  \bibinfo{year}{2022}\natexlab{}.
\newblock \showarticletitle{Air Combat Maneuver Decision Method Based on A3C
  Deep Reinforcement Learning}.
\newblock \bibinfo{journal}{\emph{Machines}} \bibinfo{volume}{10},
  \bibinfo{number}{11} (\bibinfo{year}{2022}), \bibinfo{pages}{1033}.
\newblock


\bibitem[Hessel et~al\mbox{.}(2018)]%
        {hessel2018rainbow}
\bibfield{author}{\bibinfo{person}{Matteo Hessel}, \bibinfo{person}{Joseph
  Modayil}, \bibinfo{person}{Hado van Hasselt}, \bibinfo{person}{Tom Schaul},
  \bibinfo{person}{Georg Ostrovski}, \bibinfo{person}{Will Dabney},
  \bibinfo{person}{Dan Horgan}, \bibinfo{person}{Bilal Piot},
  \bibinfo{person}{Mohammad~Gheshlaghi Azar}, {and} \bibinfo{person}{David
  Silver}.} \bibinfo{year}{2018}\natexlab{}.
\newblock \showarticletitle{Rainbow: Combining Improvements in Deep
  Reinforcement Learning}. In \bibinfo{booktitle}{\emph{Proceedings of the AAAI
  Conference on Artificial Intelligence}}, Vol.~\bibinfo{volume}{32}.
  \bibinfo{pages}{3150--3157}.
\newblock


\bibitem[Higby and Col(2005)]%
        {higby2005promise}
\bibfield{author}{\bibinfo{person}{Lt~Col~Patrick Higby} {and}
  \bibinfo{person}{USAF Col}.} \bibinfo{year}{2005}\natexlab{}.
\newblock \showarticletitle{Promise and reality: Beyond visual range (BVR)
  air-to-air combat}.
\newblock \bibinfo{journal}{\emph{Air War College (AWC) Electives Program: Air
  Power Theory, Doctrine, and Strategy: 1945--Present}}  \bibinfo{volume}{30}
  (\bibinfo{year}{2005}).
\newblock


\bibitem[Hu et~al\mbox{.}(2021)]%
        {hu2021application}
\bibfield{author}{\bibinfo{person}{Dongyuan Hu}, \bibinfo{person}{Rennong
  Yang}, \bibinfo{person}{Jialiang Zuo}, \bibinfo{person}{Ze Zhang},
  \bibinfo{person}{Jun Wu}, {and} \bibinfo{person}{Ying Wang}.}
  \bibinfo{year}{2021}\natexlab{}.
\newblock \showarticletitle{Application of deep reinforcement learning in
  maneuver planning of beyond-visual-range air combat}.
\newblock \bibinfo{journal}{\emph{IEEE Access}}  \bibinfo{volume}{9}
  (\bibinfo{year}{2021}), \bibinfo{pages}{32282--32297}.
\newblock


\bibitem[Hu et~al\mbox{.}(2022)]%
        {hu2022autonomous}
\bibfield{author}{\bibinfo{person}{Jinwen Hu}, \bibinfo{person}{Luhe Wang},
  \bibinfo{person}{Tianmi Hu}, \bibinfo{person}{Chubing Guo}, {and}
  \bibinfo{person}{Yanxiong Wang}.} \bibinfo{year}{2022}\natexlab{}.
\newblock \showarticletitle{Autonomous maneuver decision making of dual-UAV
  cooperative air combat based on deep reinforcement learning}.
\newblock \bibinfo{journal}{\emph{Electronics}} \bibinfo{volume}{11},
  \bibinfo{number}{3} (\bibinfo{year}{2022}), \bibinfo{pages}{467}.
\newblock


\bibitem[Liu et~al\mbox{.}(2022)]%
        {liu2022multi}
\bibfield{author}{\bibinfo{person}{Xiaoxiong Liu}, \bibinfo{person}{Yi Yin},
  \bibinfo{person}{Yuzhan Su}, {and} \bibinfo{person}{Ruichen Ming}.}
  \bibinfo{year}{2022}\natexlab{}.
\newblock \showarticletitle{A Multi-UCAV Cooperative Decision-Making Method
  Based on an MAPPO Algorithm for Beyond-Visual-Range Air Combat}.
\newblock \bibinfo{journal}{\emph{Aerospace}} \bibinfo{volume}{9},
  \bibinfo{number}{10} (\bibinfo{year}{2022}), \bibinfo{pages}{563}.
\newblock


\bibitem[Mnih et~al\mbox{.}(2015)]%
        {mnih2015human}
\bibfield{author}{\bibinfo{person}{Volodymyr Mnih}, \bibinfo{person}{Koray
  Kavukcuoglu}, \bibinfo{person}{David Silver}, \bibinfo{person}{Andrei~A
  Rusu}, \bibinfo{person}{Joel Veness}, \bibinfo{person}{Marc~G Bellemare},
  \bibinfo{person}{Alex Graves}, \bibinfo{person}{Martin Riedmiller},
  \bibinfo{person}{Andreas~K Fidjeland}, \bibinfo{person}{Georg Ostrovski},
  {et~al\mbox{.}}} \bibinfo{year}{2015}\natexlab{}.
\newblock \showarticletitle{Human-level control through deep reinforcement
  learning}.
\newblock \bibinfo{journal}{\emph{Nature}} \bibinfo{volume}{518},
  \bibinfo{number}{7540} (\bibinfo{year}{2015}), \bibinfo{pages}{529--533}.
\newblock


\bibitem[Piao et~al\mbox{.}(2020)]%
        {piao2020beyond}
\bibfield{author}{\bibinfo{person}{Haiyin Piao}, \bibinfo{person}{Zhixiao Sun},
  \bibinfo{person}{Guanglei Meng}, \bibinfo{person}{Hechang Chen},
  \bibinfo{person}{Bohao Qu}, \bibinfo{person}{Kuijun Lang},
  \bibinfo{person}{Yang Sun}, \bibinfo{person}{Shengqi Yang}, {and}
  \bibinfo{person}{Xuanqi Peng}.} \bibinfo{year}{2020}\natexlab{}.
\newblock \showarticletitle{Beyond-visual-range air combat tactics
  auto-generation by reinforcement learning}. In \bibinfo{booktitle}{\emph{2020
  International Joint Conference on Neural Networks (IJCNN)}}. IEEE,
  \bibinfo{pages}{1--8}.
\newblock


\bibitem[Schaul et~al\mbox{.}(2015)]%
        {schaul2015prioritized}
\bibfield{author}{\bibinfo{person}{Tom Schaul}, \bibinfo{person}{John Quan},
  \bibinfo{person}{Ioannis Antonoglou}, {and} \bibinfo{person}{David Silver}.}
  \bibinfo{year}{2015}\natexlab{}.
\newblock \showarticletitle{Prioritized Experience Replay}.
\newblock \bibinfo{journal}{\emph{arXiv preprint arXiv:1511.05952}}
  (\bibinfo{year}{2015}).
\newblock


\bibitem[Wang et~al\mbox{.}(2015)]%
        {wang2015dueling}
\bibfield{author}{\bibinfo{person}{Ziyu Wang}, \bibinfo{person}{Tom Schaul},
  \bibinfo{person}{Matteo Hessel}, \bibinfo{person}{Hado van Hasselt},
  \bibinfo{person}{Marc Lanctot}, {and} \bibinfo{person}{Nando de Freitas}.}
  \bibinfo{year}{2015}\natexlab{}.
\newblock \showarticletitle{Dueling network architectures for deep
  reinforcement learning}. In \bibinfo{booktitle}{\emph{International
  Conference on Machine Learning}}. PMLR, \bibinfo{pages}{1995--2003}.
\newblock


\bibitem[Zhang et~al\mbox{.}(2022a)]%
        {zhang2022maneuver}
\bibfield{author}{\bibinfo{person}{Hongpeng Zhang}, \bibinfo{person}{Yujie
  Wei}, \bibinfo{person}{Huan Zhou}, {and} \bibinfo{person}{Changqiang Huang}.}
  \bibinfo{year}{2022}\natexlab{a}.
\newblock \showarticletitle{Maneuver Decision-Making for Autonomous Air Combat
  Based on FRE-PPO}.
\newblock \bibinfo{journal}{\emph{Applied Sciences}} \bibinfo{volume}{12},
  \bibinfo{number}{20} (\bibinfo{year}{2022}), \bibinfo{pages}{10230}.
\newblock


\bibitem[Zhang et~al\mbox{.}(2022b)]%
        {zhang2022autonomous}
\bibfield{author}{\bibinfo{person}{Hongpeng Zhang}, \bibinfo{person}{Huan
  Zhou}, \bibinfo{person}{Yujie Wei}, {and} \bibinfo{person}{Changqiang
  Huang}.} \bibinfo{year}{2022}\natexlab{b}.
\newblock \showarticletitle{Autonomous maneuver decision-making method based on
  reinforcement learning and Monte Carlo tree search}.
\newblock \bibinfo{journal}{\emph{Frontiers in Neurorobotics}}
  (\bibinfo{year}{2022}).
\newblock


\end{thebibliography}

\end{document}